\pdfoutput=1

\documentclass[11pt]{article}

\usepackage[]{acl}

\usepackage{times}
\usepackage{latexsym}

\usepackage[T1]{fontenc}
\usepackage[utf8]{inputenc}

\usepackage{microtype}
\usepackage{graphicx} 
\usepackage{xspace}
\usepackage{amsfonts}
\usepackage{subfigure}
\usepackage{amsmath}
\usepackage{mathtools}
\usepackage{arydshln} 

\newcommand{\bert}{BERT}
\newcommand{\xlnet}{XLNet}  
\newcommand{\roberta}{RoBERTa}

\newcommand{\sentix}{SentiX}
\newcommand{\sentilare}{SentiLARE}  
\newcommand{\skep}{SKEP}

\newcommand{\ourmodel}{KESA}

\newcommand{\mr}{{MR}}
\newcommand{\ssttwo}{{SST2}}
\newcommand{\sstfive}{{SST5}}
\newcommand{\imdbtwo}{{IMDB}}

\newcommand{\swc}{{sentiment word selection}\xspace}
\newcommand{\csp}{{conditional sentiment prediction}\xspace}

\title{\ourmodel: A Knowledge Enhanced Approach To Sentiment Analysis}

\author{Qinghua Zhao, Shuai Ma, Shuo Ren \\
        SKLSDE Lab, Beihang University, Beijing, China\\
        \{zhaoqh, mashuai, shuoren\}@buaa.edu.cn}

\begin{document}
\maketitle

\begin{abstract}
Though some recent works focus on injecting sentiment knowledge into pre-trained language models, they usually design mask and reconstruction tasks in the post-training phase. This paper aims to integrate sentiment knowledge in the fine-tuning stage. To achieve this goal, we propose two sentiment-aware auxiliary tasks named \swc and \csp and, correspondingly, integrate them into the objective of the downstream task. The first task learns to select the correct sentiment words from the given options. The second task predicts the overall sentiment polarity, with the sentiment polarity of the word given as prior knowledge. In addition, two label combination methods are investigated to unify multiple types of labels in each auxiliary task. Experimental results demonstrate that our approach consistently outperforms baselines (achieving a new state-of-the-art) and is complementary to existing sentiment-enhanced post-trained models. 
The codes are released at \url{https://github.com/lshowway/KESA}.
\end{abstract}

\section{Introduction}
Sentence-level sentiment analysis aims to classify the overall sentiment of a sentence, which has received considerable attention in natural language processing ~\citep{liu2012sentiment, zhang2018deep, zhang2022towards}. 
Recently, pre-trained language models (PLMs) have achieved state-of-the-art (SOTA) performance on many natural language processing (NLP) tasks, including sentiment analysis. 
However, it is still challenging to integrate external knowledge into PLMs~\citep{lei2018multi, xu2019bert, liu2020k, wei2021knowledge, yang2021survey, cui2021knowledge, zhang-etal-2022-incorporating}. 

Recently, sentiment dictionary, a commonly used sentiment knowledge, has been injected into PLMs~\citep{wu-etal-2022-sentiment}. A common practice is to post-train~\cite{xu-etal-2019-bert}, i.e., continue pre-training, self-designed tasks on domain-specific corpora. These tasks include sentiment word prediction task, word sentiment prediction task, or aspect-sentiment pairs prediction~\citep{xu2019bert, tian2020skep, ke-etal-2020-sentilare, gururangan2020don, gu2020train, tian-etal-2021-enhancing, li2021sentiprompt}, just to name a few.
Specifically, they are usually designed according to the paradigm of the mask language model (MLM), where sentiment words are first masked and then recovered  (including their polarities) in the output layer. 
Though effective, we argue that these methods can be further boosted by directly injecting sentiment knowledge, e.g., sentiment polarity, into the output layer when fine-tuning the downstream tasks.

In this paper, we aim to inject sentiment knowledge into the fine-tuning phase directly, making it complementary to existing methods. 
For this aim, we propose two novel auxiliary tasks.
The first task is \swc (SWS), aiming to select the sentiment words that belong to the input from the given options, which comprises of $K+1$ options (i.e., one ground-truth and $K$ negative words). 
The second task is \csp (CSP), which pushes the model to predict the sentence polarity (i.e., sentiment), with the word (within the sentence) polarity given as prior information. 
It can be seen as a simplified main task (i.e., sentence-level sentiment analysis).
Different from existing sentiment polarity prediction task, CSP treats the word sentiment (extracted from the sentiment dictionary) as prior information at the input end instead of as the ground-truth label at the output end.
Intuitively, this transformation can reduce the dependency on the quality of the sentiment dictionary. Otherwise, though effective, its interpretability will be impaired.
Besides, since more than one type of label (e.g., sentence/word polarity label) is included, two label combination methods, i.e., the joint combination and the conditional combination, are therefore investigated.
We are the first (earlier than ~\citep{zhang-etal-2022-incorporating}) to inject sentiment knowledge in the fine-tuning stage.
Our method starts by building the sentiment dictionary out of public resources and recognizing all the sentiment words in the input sentence. Next, each auxiliary task is added to the task-specific (i.e., output) layer. 
Finally, the auxiliary loss is added to the main loss to achieve the total loss.

We conduct experiments to demonstrate the further effectiveness of our proposed approach, and run ablation studies to verify the effectiveness of each auxiliary task. 
Analysis studies are also performed to compare the impacts of hyper-parameters or modules.
With \ourmodel, the performance further outperforms the state-of-the-art by (0.76\%, 0.75\%) accuracy on \mr~and \sstfive, respectively.

\section{Related Work}\label{related_work}
\textbf{Pre-training Language Models.}
Pre-trained language models have achieved remarkable improvements in many NLP tasks, and many variants of PLMs have been proposed. 
For example, GPT, GPT-2 and GPT-3~\citep{radford2018improving, radford2019language, brown2020language},
BERT~\citep{devlin2019bert}, XLNet~\citep{yang2019xlnet} and ALBERT~\citep{lan2019albert},
ERNIE~\citep{sun2020ernie}, 
BART~\citep{lewis2020bart} and RoBERTa~\citep{liu2019roberta}. 
Most PLMs are pre-trained on large-scale unlabeled general corpora by pre-training tasks, pushing models to pay attention to deeper semantic information.
Currently, PLMs are the fundamental models across NLP tasks, and the pre-training tasks mentioned above are summarized in Table~\ref{Pretext_tasks}.

\textbf{Knowledge Enhanced Post-trained Language Models.}
\begin{table}[t]
\centering
\resizebox{.90\columnwidth}{!}{
\begin{tabular}{ll}
\hline \textbf{Model} & \textbf{Pre/Post-training Tasks}\\ \hline
\bert& MLM and NSP \\
ALBERT&sentence order prediction \\
ERNIE&knowledge mask\\
&     sentence reordering \\
BART&token mask/deletion\\
&    sentence permutation \\
\hline
\skep& sentiment word prediction\\
&     word polarity prediction\\
&     aspect-sentiment pair prediction \\
\sentilare& sentiment word prediction\\
&           word polarity prediction\\
&           POS label prediction\\
&           joint prediction \\
\sentix& sentiment word prediction\\
&     word polarity prediction\\
&   emotion prediction\\
&      rating prediction\\
\hline
\textbf{\ourmodel}& \swc\\
&   \csp\\
\hline
\end{tabular}}
\caption{\label{Pretext_tasks} An overview of tasks. The first block is pre-training tasks, and the second block is knowledge-related tasks. NSP  refers to the next sentence prediction. }
\end{table}
\begin{figure*}[t]
\centering
\includegraphics[width=0.8\textwidth]{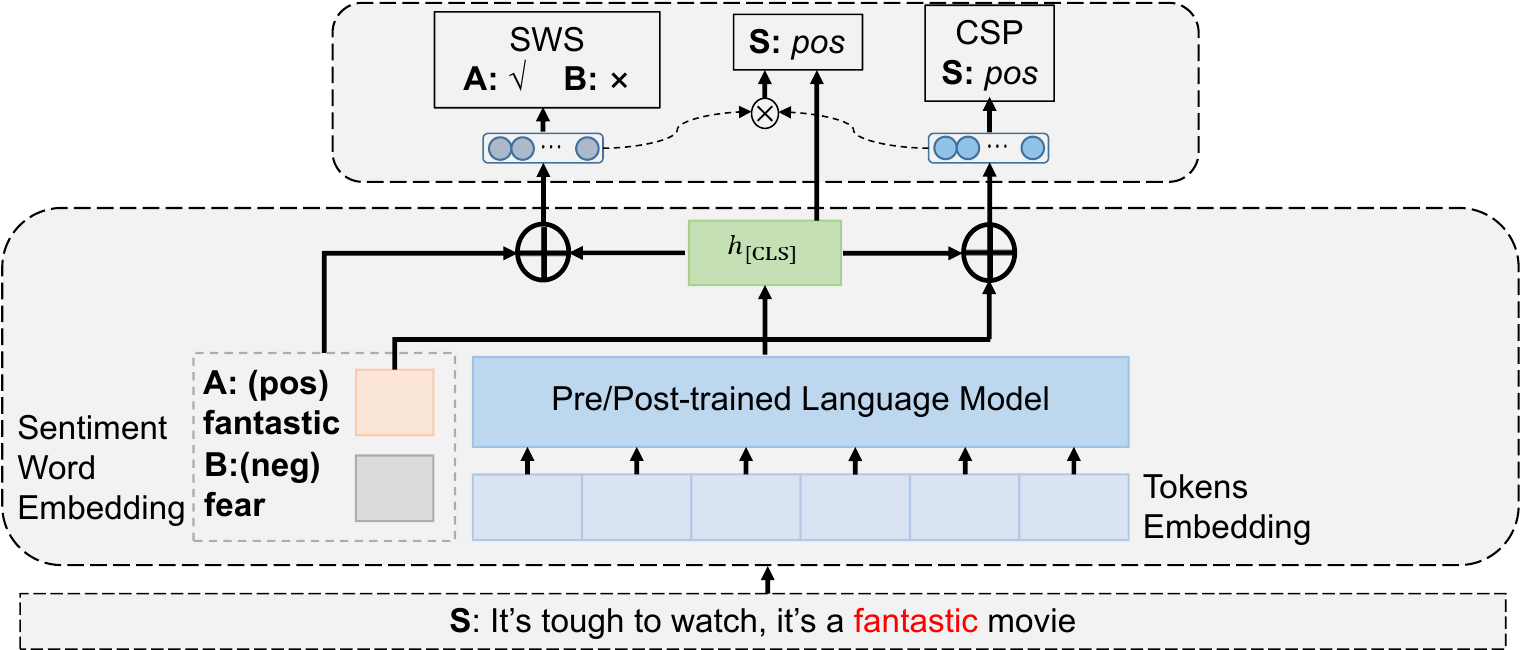}
\caption{Overview of \ourmodel. 
Firstly, at the bottom of this figure, the sentence $S$ is tokenized into subwords and input into PLMs to obtain context representation $h_{[\text{CLS}]}$. Meanwhile, sentiment word \texttt{fantastic} and its sentiment \texttt{positive} are recognized by external sentiment dictionary and a sentiment word \texttt{fear} is randomly selected from the sentiment dictionary. 
Secondly, for the sentiment word selection task, \texttt{fantastic} and \texttt{fear} are treated as options. For the conditional sentiment prediction task, only the ground-truth sentiment word \texttt{fantastic} and its corresponding sentiment  \texttt{positive} are considered. 
}
\label{framework}
\end{figure*}
External knowledge, including linguistic knowledge (e.g., part-of-speech, hyponym and synonym), factual knowledge (including items from Wikidata~\citep{DennyVrandecic2012WikidataAN}, ConceptNet~\citep{RobertSpeer2016ConceptNet5A} and Wikipedia) or domain-specific knowledge (e.g., sentiment polarity), can boost the generalization abilities of PLMs~\citep{yin2022survey,kgqgipm24}.
Several works have recently attempted injecting knowledge into PLMs, where the input format or model structure is modified, and knowledge-aware tasks are designed~\cite{zhang2019ernie, sun2021ernie, 10.1145/3383313.3412237, su-etal-2021-enhanced, cui2021knowledge, yu2022dict, yu2022jaket,agentqgcikm24,agenteaicde25,hydra26}. For example, 
ERNIE 3.0~\citep{sun2021ernie} appends triples, e.g., (Andersen, Write, Nightingale), ahead of the original input sentence, and designs tasks to predict the relation "Write" in the triple.
K-BERT~\cite{liu2020k} appends triples as branches to each entity (when fine-tuning downstream tasks) involved in the input sentence to form a sentence tree. 
K-Adapter~\citep{wang-etal-2021-k} designs adapters and regards them as a plug-in with knowledge representations. These adapters are decoupled from the backbone PLMs and pre-trained from scratch by self-designed tasks, e.g., predicting relations in triples and labels of dependency parser.
\citep{cui2021knowledge} also considers adding two auxiliary training objectives when fine-tuning the dialogue generation task, including conjecturing the meaning of the masked entity and predicting its hypernym. Different from ours, it is also designed according to the paradigm of MLM (i.e., masking entities and predicting their associated attributes in the knowledge base).

\textbf{Knowledge Enhanced Post-trained Language Models for Sentiment Analysis.}
Some domain-specific knowledge (including sentiment dictionary) is used for the sentiment analysis task. Generally, these methods inject sentiment-related information into PLMs by designing sentiment-aware tasks and then post-train them on large-scale domain-specific corpora~\citep{tian2020skep, ke-etal-2020-sentilare, zhou2020sentix, tian-etal-2021-enhancing}. 
For example, 
\skep~\citep{tian2020skep} designs sentiment word prediction, word polarity prediction, and aspect-sentiment pair prediction task to enhance PLMs with sentiment words. 
\sentilare~\citep{ke-etal-2020-sentilare} also designs sentiment word prediction, word polarity prediction, and joint prediction tasks.
\sentix~\citep{zhou2020sentix} designs sentiment word prediction, word polarity prediction, emoticon and rating prediction tasks. Table~\ref{Pretext_tasks} summarizes the tasks mentioned above.
Like MLM, they mask sentiment words in the input and then recover their related sentiment information in the output. 
~\citep{tian-etal-2021-enhancing} associates each aspect term with its corresponding dependency relation types as knowledge to enhance aspect-level sentiment analysis. 
~\citep{li2021sentiprompt} enhances aspects and opinions with sentiment knowledge enhanced prompts.
Besides, \citep{zhang-etal-2022-incorporating}\footnote{We do not take it as a baseline as it is designed for aspect-base sentiment analysis task.} also injects sentiment knowledge in the fine-tuning phase, it incorporates and updates a lightweight dynamic reweighting adapter when fine-tuning the downstream tasks (we are earlier than this).
Our work is different from the above. We propose two novel auxiliary objectives and integrate them with the main objective when fine-tuning the downstream tasks. Furthermore, instead of treating word polarity as a ground-truth label, we treat it as prior knowledge to assist in predicting the overall sentiment. 
We also investigate two label combination methods to consider several types of labels simultaneously.

\section{Methodology}

Figure~\ref{framework} illustrates the framework of \ourmodel.
In order to integrate sentiment-related information when fine-tuning the downstream tasks, we propose two straightforward auxiliary tasks. 
The subsequent subsections will detail the two proposed auxiliary tasks (Section~\ref{subtask_A} and \ref{subtask_B}), two label combination methods (including joint and conditional combination, Section~\ref{label_combination}) and a weighted loss function (Section~\ref{loss_function}) .
For convenience, we first give some notations used in the following subsections.

Formally, $L = \{l_1, l_2, \cdots, l_M\}$  denotes the sentiment dictionary with the size of $M$ (i.e., including $M$ sentiment words), and $S = \{w_1, w_2, \cdots, w_N\}$ denotes an input sentence of length $N$. 
$P_S \in C$ and $P_w \in Z$ denote the polarity of the sentence $S$ and the word $w$, respectively, where $C$ is the sentence sentiment polarity label set, and $Z$ is the word sentiment label set. 
$Y \in \{0, 1\}$ denotes the ascription relationship label set between the word and the sentence,  e.g., $Y_{w, S} = 1$ means that the sentiment word $w$ belongs to the sentence $S$. 
$d$ is the dimension of embeddings.

\subsection{Main Task}
The main task, i.e., sentence-level sentiment analysis, is to predict the sentiment label $P_S$ given the input sentence $S$. Firstly, the input $S$ is passed through PLMs to get the context representation $h_{[\text{CLS}]}$. Then the context representation is fed into a linear layer and a Softmax layer to get the probability $\hat{P}_S$ over sentiment label set, i.e., $\hat{P}_S = \text{Softmax}(W_1 h_{[\text{CLS}]} + b_1)$, where $W_1$ and $b_1$ are the model parameters.

\subsection{Task A: Sentiment Word Selection}\label{subtask_A}
Existing sentiment word prediction tasks usually randomly mask some identified sentiment words in the input, and then predict them in the output layer (in the pre/post-training phase) by computing the probability distribution over the vocabulary of sentiment words. 
Compared with the number of classes ($\left| C \right|$) of the downstream task, the sentiment word vocabulary size is much larger and directly transferring the above method to the fine-tuning stage may push PLMs to focus on more complex tasks, i.e., the auxiliary tasks.
To avoid this issue, we design the \swc (SWS) task to require PLMs to select the ground-truth sentiment word from given options.

\begin{figure}[t]
\centering
\includegraphics[width=\linewidth]{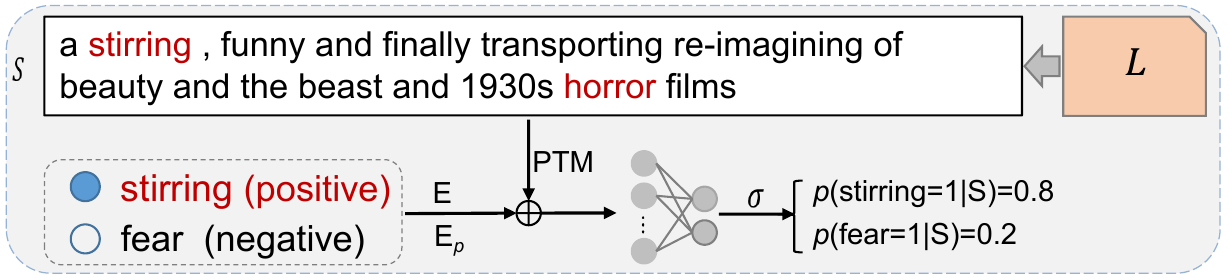}
\caption{A demonstration of auxiliary task A. The sentence is sampled from \ssttwo~dataset,   $\sigma$ refers to the Softmax layer. It shows that given sentence $\text{S}$, two sentiment word options (i.e., ``stirring'' and ``fear'') and their associated sentiment polarities (``positive'' and ``negative''), ``stirring'' has more probability of being in $\text{S}$.}
\label{A_example}
\end{figure}

Given a training sample $(S, P_S)$, we first recognize all the sentiment words in $S$ according to the sentiment dictionary $L$ by exact word match. Then, we randomly choose one sentiment word $w_i$ (i.e., positive option) from them and record its sentiment polarity as $P_{w_i}$. 
Meanwhile, we randomly sample one sentiment word from $L$ as $w_j$ (i.e., negative option) and record its sentiment polarity as $P_{w_j}$ ($w_j \neq w_i$).
Next, we tokenize $S$ into a subwords sequence, add ``$\text{[CLS]}$'' ahead of the sequence, lookup each subword embedding and input them into PLMs. The first token representation ($h_{[\text{CLS}]}$) of the last layer of PLMs is treated as the context representation (from the view of the representations of sentiment word options).

Meanwhile, we extract the embeddings of the sentiment word options $w_i$, $w_j$ as $e_i$ and $e_j$, and the embeddings of its sentiment polarity $p_{w_i}$, $p_{w_j}$ as $e'_i$ and $e'_j$, respectively.
For each option, we add the context representation, word and its polarity embedding together, and then input them into a linear layer and a Softmax layer to compute the probability $\hat{O}_A = \{ \hat{o}_i, \hat{o}_j\}$ over the given options,
\begin{equation}
    \hat{o}_x = \text{Softmax}(W_x (h_{\text{[CLS]}} + e_x + e'_x)), x\in \{i, j\}
\label{swc}
\end{equation}
$b_x$ is omitted in Eq.~\ref{swc}, and $W_x, x \in \{i, j\}$ refers to model parameters.

Figure~\ref{A_example}  gives an example of the procedure of SWS. In this example, ``stirring'', ``funny'', ``beauty'' and ``horror'' are first recognized as sentiment words.  ``stirring'' is then randomly selected as the positive option, and ``fear'' is randomly sampled as a negative option.
The sentence $S$ is input into PLMs to get the context representation $h_{\text{[CLS]}}$.
Meanwhile, the word embeddings of ``stirring'' and ``fear'' are lookup from the sentiment word embedding table $E \in \mathbb{R}^{|V_1| \times d}$, where $V_1$ refers to sentiment word vocabulary.
Correspondingly, their polarity embeddings are looked up from polarity embedding table $E_p \in \mathbb{R}^{|Z| \times d}$. $E$ and $E_p$ can be initialized from scratch and updated during the training, or cached pre-trained embeddings and frozen during the training.
Subsequently, $h_{\text{[CLS]}}$ is added to the word and polarity embeddings of the positive (or negative) option, to produce sentiment-enhanced (or polluted) context representation, which is then used to compute the probability of being the ground-truth.

\subsection{Task B: Conditional Sentiment Prediction}\label{subtask_B}
\begin{figure}[tp]
\centering
\includegraphics[width=\linewidth]{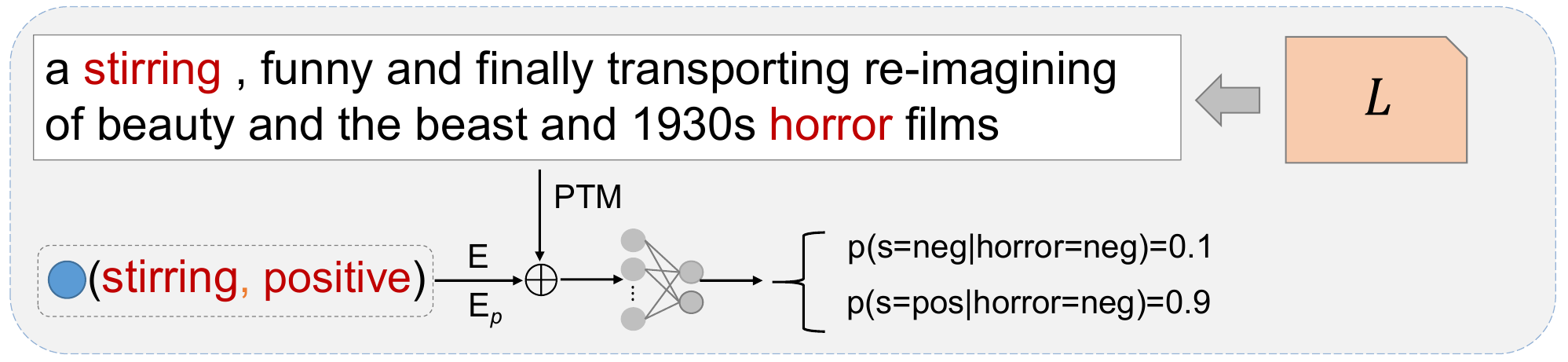}
\caption{A demonstration of auxiliary task B. This sample shows that the sentiment word, i.e., ``horror'' and its polarity (``negative'') is given as prior knowledge.}
\label{B_example}
\end{figure}

Existing word polarity prediction tasks usually replace sentiment words with ``\texttt{[MASK]}'' in the input, and recover their sentiment labels in the output layer (in the post-training stage). In this process, sentiment words and their sentiment labels are extracted by sentiment dictionary or statistical methods, which may be inaccurate. Though effective, we argue there are still challenges in interpretability, since it is hard to discriminate which (domain corpus or sentiment-aware tasks) boosts the performance.
To avoid the negative impacts of inaccurate polarity of sentiment words, we design the \csp task, which treats the polarity of sentiment words as prior information instead of the ground-truth label.

More specifically, given a training sample $(S, P_S)$, similar to SWS, we first choose one sentiment word $w_i$ (i.e., positive option detailed in Section~\ref{subtask_A}) from all recognized sentiment words in $S$, meanwhile recording its sentiment polarity $P_{w_i}$, sentiment word embedding $e_i$ and its polarity embedding $e'_{i}$.
Next the sentence $S$ is fed into PLMs to get the context representation $h_{\text{{[CLS]}}}$. Afterwards, we add $e_i$ and $e'_{i}$ to $h_{\text{[CLS]}}$ to create sentiment-enhanced context representation,  then passing them through a linear layer and a Softmax layer to predict the probability distribution over sentence sentiment label set $C$, i.e.,
\begin{equation}
    \hat{O}_B = \text{Softmax}(W_3 (h_{\text{[CLS]}} + e_i + e'_{i}) + b_3)
\label{csp}
\end{equation}
where $W_3, b_3$ are model parameters.
CSP learns the influence of a word polarity on the polarity of its assigned sentence.
In a broader sense, how local information affects global information.
Figure~\ref{B_example} gives an example of the auxiliary task B.

\subsection{Label Combination}\label{label_combination}
For each auxiliary task, we need to unify all kinds of labels. 
To be specific, for the SWS task, in addition to the sentence polarity label $P_S$, we also need to consider the word ascription label $Y$. Correspondingly, for the CSP task, sentence polarity $P_S$ and word polarity $P_{w}$ are both involved.
Intuitively, multiple kinds of labels can describe the input sentence from different perspectives, and encourage the model to leverage different helpful information simultaneously~\citep{caruana1997multitask}.
To treat the involved label types in a unified manner, we explore two types of combination methods. 
The first one is joint combination, which models the joint probability distribution of the multiple kinds of labels. This method treats all kinds of labels as a single label defined on the Cartesian product of different labels. 
The second way is a conditional combination motivated by ~\citet{lee2020self}, which models the conditional probability distribution of multiple kinds of labels, predicting one kind of label with other kinds of labels as prior conditions.

\textbf{Joint combination.}
For task A (SWS), given the overall logits $\hat{O}_A$,
we need to predict the joint probability distribution of the word ascription label and the sentence polarity label. That is, $p(Y,P_S|\hat{O}_A) \in \mathbb{R}^{|Y| \times|C|}$, where $|Y|$ means the size of label set $Y$ ($\{0, 1\}$) and $|C|$ means the size of label set $P_S$. 
For task B (CSP), given the overall logits $\hat{O}_B$ in Eq.~\ref{csp}.
We predict the joint distribution of the word polarity label and the sentence polarity label.
That is, $p(P_{w},P_S|\hat{O}_B) \in \mathbb{R}^{|Z| \times|C|}$, where $|Z|$ means the number of $P_{w}$'s labels (i.e., \{\texttt{positive}, \texttt{negative}\} in our experiment).

\textbf{Conditional combination.}
For task A, given the overall logits $\hat{O}_A$, we predict the probability to sentence polarity under the condition that the word ascription label is known, i.e., $p(P_S|\hat{O}_A, Y) \in \mathbb{R}^{|C|}$.
To get this, we simply choose the according logits indexed by $Y$ from $\hat{O}_A$ followed by normalization. 
Similarly, For task B, given the overall logits $\hat{O}_B$ in Eq.~\ref{csp}, the conditional probability of sentence sentiment polarity given the word sentiment polarity is $p(P_S|\hat{O}_B,P_{w}) \in \mathbb{R}^{|C|}$.
For that, we just select the according logits indexed by $P_w$ from $\hat{O}_B$.

\subsection{Loss Function}\label{loss_function}
We take cross-entropy loss as our loss function. The loss function is defined as the cross-entropy between the predicted probability (e.g., $\hat{P}_S$, $\hat{O}_A$ and $\hat{O}_B$) and the ground-truth label $P_S$.

The loss function of the main task is:
\begin{equation}
    \mathcal{L}_{main} = - \frac{1}{|C|} \sum_{i \in C} P_S \cdot \text{log}(\hat{P}_S)
\label{main_loss}
\end{equation}

The loss function of the auxiliary tasks $\mathcal{L}_{aux}$ has the same formulation as Eq.~\ref{main_loss}, except that the predicted probability is a weighted sum of $\hat{O}_A, \hat{O}_B$: 
\begin{equation}
    W_4(p(P_S | \hat{O}_A, Y) \ || \  p(P_S | \hat{O}_B, P_w)) \in \mathbb{R}^C
\label{aux_loss}
\end{equation}
where $W_4 \in \mathbb{R}^{2 \times 1}$ is model parameters, $||$ refers to concatenation,
Note that, we omit the bias $b_4$ in Eq.~\ref{aux_loss}.
The final loss is a weighted sum,   
\begin{equation}
\label{total_loss}
    \mathcal{L} = \mathcal{L}_{main} + \gamma \mathcal{L}_{aux}
\end{equation}
where $\gamma$ is loss balance weight and $\gamma \in (0.0, 1.0)$. 
Notably, the weight of $\mathcal{L}_{main}$ is set to 1.0. We set $\gamma>0.0$ to ensure that the parameters of the auxiliary tasks can be optimized by backpropagation, and set $\gamma<1.0$ to prevent the final loss is dominated by the auxiliary task loss and diminishing the performance of the main task~\citep{liu2019loss}.


\section{Experimental Setup}

\subsection{Datasets}
Four commonly used public sentence-level sentiment analysis datasets are used for the experiment, as shown in Table ~\ref{data_statistics}. The datasets include Movie Review (MR) \cite{pang2005seeing}, Stanford Sentiment Treebank (\ssttwo~and \sstfive) \cite{socher2013recursive} and \imdbtwo. For \mr~and \imdbtwo, we adopt the data split in SentiLARE \cite{ke-etal-2020-sentilare}, due to the lack of test data in the original dataset. We evaluate the model performance in terms of accuracy.
\begin{table}
\centering
\resizebox{.90\columnwidth}{!}{
\begin{tabular}{l|l|l|c}
\hline \textbf{Dataset} & \textbf{\#Train/Valid/Test}& \textbf{\#W} & \textbf{\#C}\\ \hline
\mr& 8,534/1,078/1,050& 22&  2   \\
\ssttwo&  6,920/872/1,821& 20&  2   \\
\sstfive&  8,544/1,101/2,210&  20&  5   \\
\imdbtwo&   22,500/2,500/25,000& 280& 2 \\
\hline
\end{tabular}
}
\caption{Datasets statistics. The columns are the amount of training/validation/test sets, the average sentence length, and the number of classes, respectively.}
\label{data_statistics} 
\end{table}

\subsection{Comparison Methods}
To demonstrate the further effectiveness of the proposed method, we test the proposed auxiliary tasks on two types of competitive baselines, including popular vanilla pre-trained models (PLMs) and sentiment knowledge enhanced post-trained models. 

\textbf{Vanilla Pre-trained Language Models.} 
We use the base version of vanilla \bert~\citep{devlin2019bert}, \xlnet~\citep{yang2019xlnet} and \roberta~\citep{liu2019roberta} as our baselines, which are the most popular PLMs. 

\textbf{Sentiment Knowledge Enhanced Post-trained Language Models.} We also use two methods focusing on leveraging sentiment knowledge as baselines, i.e., \sentilare~\citep{ke-etal-2020-sentilare} and \sentix~\citep{zhou2020sentix}.
They introduce sentiment knowledge in the pre-training stage by designing sentiment-related tasks (including sentiment word prediction and word polarity prediction task). 
They continue pre-training vanilla PLMs on million scale domain-specific corpora, i.e., Yelp Dataset Challenge 2019 (6.6 million) for \sentilare, Yelp Dataset Challenge 2019 and Amazon review dataset (240 million in total) for \sentix. In terms of PLMs, \sentilare~is post-trained on \roberta-base version while \sentix~is post-trained on \bert-base version.

\textbf{\ourmodel~(Ours).} 
We also utilize the external sentiment knowledge to enhance PLMs when fine-tuning the downstream tasks by designing two auxiliary tasks (i.e., SWS and CSP). 
\ourmodel~is a complementary method to existing models (including vanilla and knowledge-enhanced PLMs).

\subsection{Sentiment Dictionary}
To build sentiment dictionary, we extract word sentiment (i.e., polarity) from SentiWordNet 3.0~\citep{baccianella2010sentiwordnet}.
Since each word in SentiWordNet 3.0 has several usage frequency levels and is linked with different semantic and sentiment scores, we set the sentiment polarity of a word according to its most vital scores (i.e., positive or negative sentiment scores). Take ``thirsty'' for example, the polarity of the most common usage is ``positive'' (with a score of 0.25), while the polarity of the third common usage is ``negative'' (with a score of -0.375).
We, therefore, set the polarity of ``thirsty'' to ``negative'', considering it has a larger weight of ``negative''. We adopt this strategy considering a lower sentiment score often means less likely to be a sentiment word.

\subsection{Implementation Details}
We implement our model using HuggingFace's Transformers. The batch size is set to 16 and 32 for \imdbtwo~and other datasets, respectively. The learning rate is set to 2e-5 for \xlnet, \roberta~and \sentilare, and 5e-5 for \bert~and \sentix. The input and output formats are consistent with each corresponding PLM. In the meantime, the input sequence length is set to 50, 512, and 128 for \mr, \imdbtwo, and other datasets, respectively, to ensure that more than 90\% of the samples are covered.
Other hyper-parameters are kept by default.
We fine-tune each model for three epochs, and the best checkpoints on the development set are used for inference. As for each dataset, we run four times with different random seeds with a reproducible implementation, and the average results are reported. Moreover, to make a fair comparison, all methods use the same seeds for the same dataset. To explore the influence of auxiliary tasks on the main task, we search the loss balance weight $\gamma$ from $\{0.01, 0.1, 0.5, 1.0\}$.

\section{Experimental Results}
In this section, we will detail the overall results, and the analysis of loss balance weight, label combination and introduced extra parameters.

\subsection{Overall Results}

\begin{table}[t]
\centering
\resizebox{0.93\columnwidth}{!}{
\begin{tabular}{l|l|l|l|l}
\hline 
\textbf{Model}&  \textbf{\mr}&   \textbf{\ssttwo}&   \textbf{\sstfive}&   \textbf{\imdbtwo}\\ 
\hline
\bert$^*$&  86.62&  91.38&  53.52&  93.45   \\
\xlnet$^*$& 88.83&  92.75&  54.95&  94.99\\
\roberta$^*$&   89.84&  94.00&  57.09&  95.13\\
\hline
\sentix$^\#$&   $-$ &   93.30&  55.57&  94.78\\
\sentix$^*$&    86.81&  92.23&  55.59&  94.62\\
\sentilare$^\#$& 90.82&  $-$&   58.59&  95.71  \\
\sentilare$^*$& 90.50&  94.58&  58.54&  95.73\\
\hline
\ourmodel& \textbf{91.26$^\ddagger$}&  \textbf{94.98$^\ddagger$}&  \textbf{59.26$^{**}$}&  \textbf{95.83$^{**}$}\\
\hline
\end{tabular}
}
\caption{Overall accuracy (joint combination is adopted here). The marker $\#$ denotes the original reported results while $-$ means not available. The marker $*$ refers to our re-implementation. $**$ and $\ddagger$ indicate that our model significantly outperforms the best baselines with \textit{t}-test, \textit{p}-value $<$ 0.01 and 0.05, respectively.}
\label{table_overall_results}
\end{table}

Table~\ref{table_overall_results} reports the results w.r.t. the accuracy. 
Note that, we only report the results of \ourmodel~fine-tuned on the checkpoints released by \sentilare, since it performs best (others will be detailed next section).
We find that through post-training on 240 million samples, \sentix~(based on BERT-base)~shows improvements of (0.19\%, 0.85\%, 2.07\%, 1.17\%) accuracy, respectively.
Similarly, post-training on 6.6 million samples, \sentilare~(RoBERTa-base)~outperforms the comparad method by (0.66\%, 0.58\%, 1.45\%, 0.60\%), respectively. 
Based on these improvements, \ourmodel~can further improve the accuracy by (0.76\%, 0.40\%, 0.75\%, 0.10\%), demonstrating that \ourmodel~is complementary to existing sentiment-enhanced post-trained methods.

\subsection{Ablation Results}

\begin{table}[t]
\centering
\resizebox{0.93\columnwidth}{!}{
\begin{tabular}{l|l|l|l|l}
\hline 
\textbf{Model}&  \textbf{\mr}&   \textbf{\ssttwo}&   \textbf{\sstfive}&   \textbf{\imdbtwo}\\ 
\hline
\xlnet$^*$& 88.83&  92.75&  54.95&  94.99\\
$\Delta$+SWS&    0.22&	0.72&	0.56&	0.04\\
$\Delta$+CSP&    0.48&	0.04&	0.50&	-0.02\\
$\Delta$+\ourmodel&  0.27&	0.26&	0.99&	0.01\\
\hline

\bert$^*$&  86.62&  91.38&  53.52&  93.45   \\
$\Delta$+SWS&   -0.32&	0.08&	0.69&	0.14\\
$\Delta$+CSP&   -0.17&	0.32&	0.86&	0.06\\
$\Delta$+\ourmodel&    -0.33&	0.18&	0.61&	0.06 \\
\hline
\sentix$^*$&    86.81&  92.23&  55.59&  94.62\\
$\Delta$\sentix$^*$&    0.19&	0.85&	2.07&	1.17\\
$\Delta$+SWS&   0.50&   -0.03&	0.15&	0.09\\
$\Delta$+CSP&   0.54&	0.01&	0.24&	-0.01\\
$\Delta$+\ourmodel&     0.55&	0.29&	0.19&	-0.05\\
\hline

\roberta$^*$&   89.84&  94.00&  57.09&  95.13\\
$\Delta$+SWS&   -0.03&	0.22&	0.13&	0.27\\
$\Delta$+CSP&   0.02&	0.17&	0.15&	0.31\\
$\Delta$+\ourmodel& 0.23&	0.40&	0.09&	0.33\\
\hline
\sentilare$^*$& 90.50&  94.58&  58.54&  95.73\\
$\Delta$\sentilare$^*$&  0.66&	0.58&	1.45&	0.60\\
$\Delta$+SWS&   0.24&	0.14&	0.75&	0.07\\
$\Delta$+CSP&   0.60&	0.33&	0.05&	0.07\\
$\Delta$+\ourmodel& 0.76&	0.40&	0.72&	0.10\\

\hline
\end{tabular}
}
\caption{Ablation studies of each task. "+SWS" and "+CSP" refer to that we fine-tune the models with SWS and CSP solely, respectively.  "+\ourmodel" represents that both auxiliary tasks are adopted. The marker $*$ refers to our re-implementation.}
\label{table_ablation_results}
\end{table}

To demonstrate the individual benefits of the two auxiliary tasks to each baseline PLMs, we perform ablation experiments and tabulate the results in Table~\ref{table_ablation_results}.
Overall, \ourmodel~achieves consistent improvements over both vanilla and sentiment-enhanced PLMs. 
Adding SWS to the baseline PLMs improves accuracy by a maximum of 0.75\%, and further pushes the overall accuracy to 59.29\% (\sstfive), exceeding the previous sentiment-enhanced best of 58.54\%.
The results verify that the word ascription label pushes the model to focus more on the interactions between the sentiments of word and sentence, and this kind of interactions between sentence sentiment (can be seen as global information) and word sentiment (treated as local information) can promote the main task.
With the addition of CSP, the test set accuracy jumped 0.86\% from 53.52\% to 54.38\% (\sstfive), even improving over the previous best sentiment-enhanced result by 0.60\% (\mr).
The results demonstrate that explicitly adding the sentiment of a word brings more information and lowers the difficulty of the main task. 
Besides, we can see that integrating \ourmodel~with sentiment-enhanced PLMs obtains more gains than that with vanilla PLMs, we attribute this to that the former can achieve better semantic representation of sentiment words.
Furthermore, combining the two auxiliary tasks is not necessarily superior to sole use. It is presumably because multiple tasks may promote or compete with each other (negative learning)~\citep{bingel2017identifying}. 
Above all, these results remind us that the combinations of multiple tasks need to be carefully analyzed.
Even so, \ourmodel~still gets further improvements on all evaluated datasets in most cases.

\subsection{Analysis on Loss Balance Weight}

\begin{figure*}[ht]
\centering
\subfigure[\mr]{
\begin{minipage}[t]{0.23\linewidth}
\centering
\includegraphics[width=1.6in]{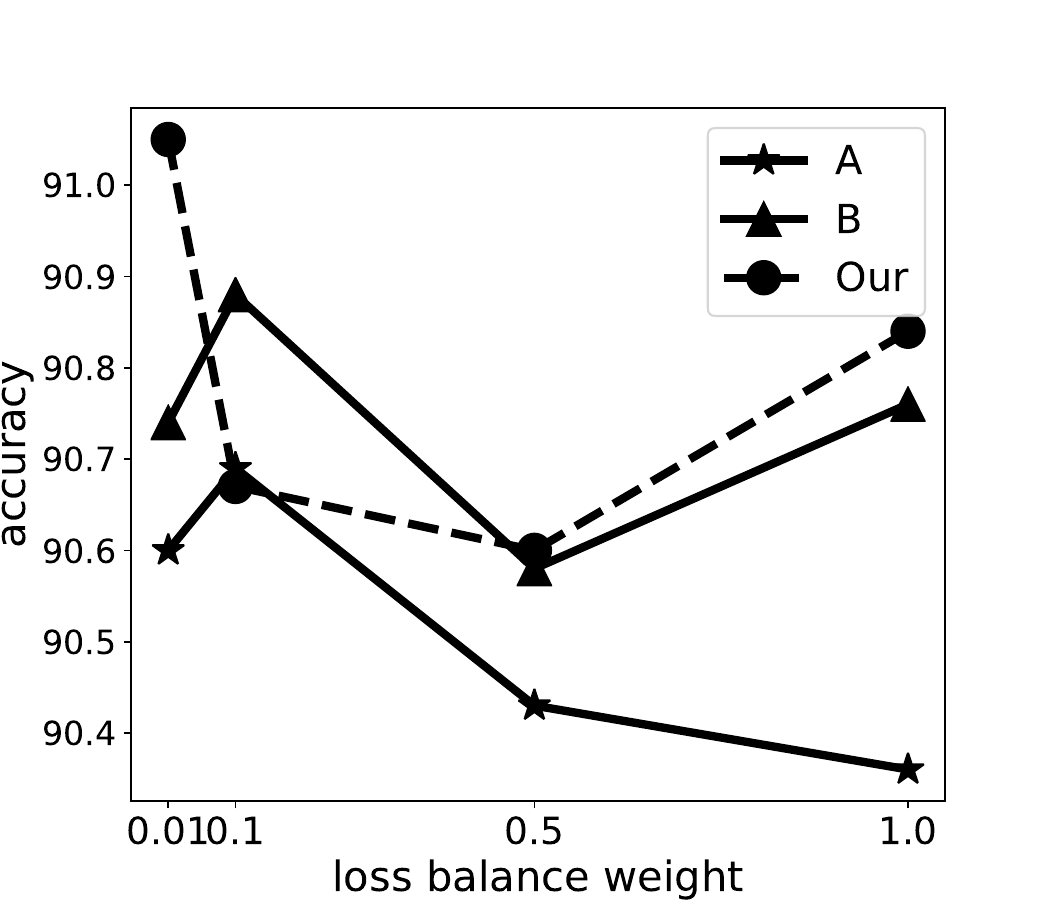}
\end{minipage}
\label{fig_a}
}
\subfigure[\ssttwo]{
\begin{minipage}[t]{0.23\linewidth}
\centering
\includegraphics[width=1.6in]{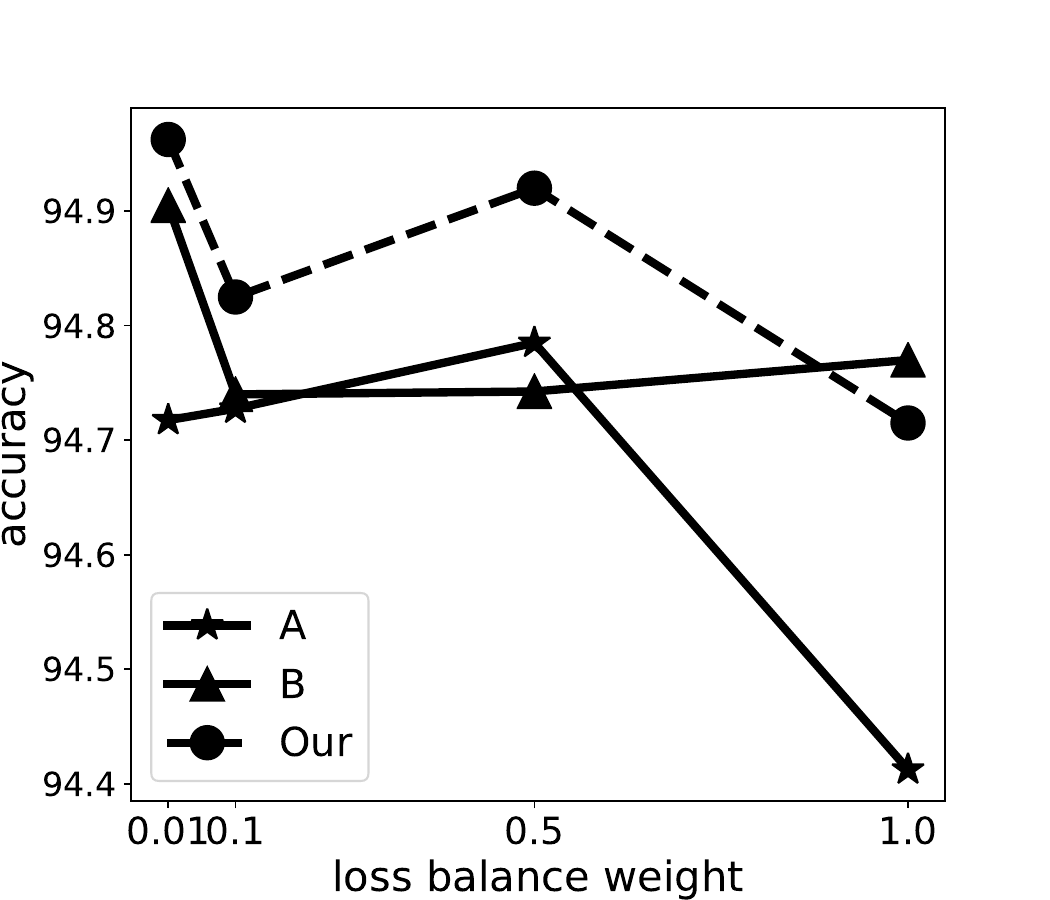}
\end{minipage}
\label{fig_b}
}
\subfigure[\sstfive]{
\begin{minipage}[t]{0.23\linewidth}
\centering
\includegraphics[width=1.6in]{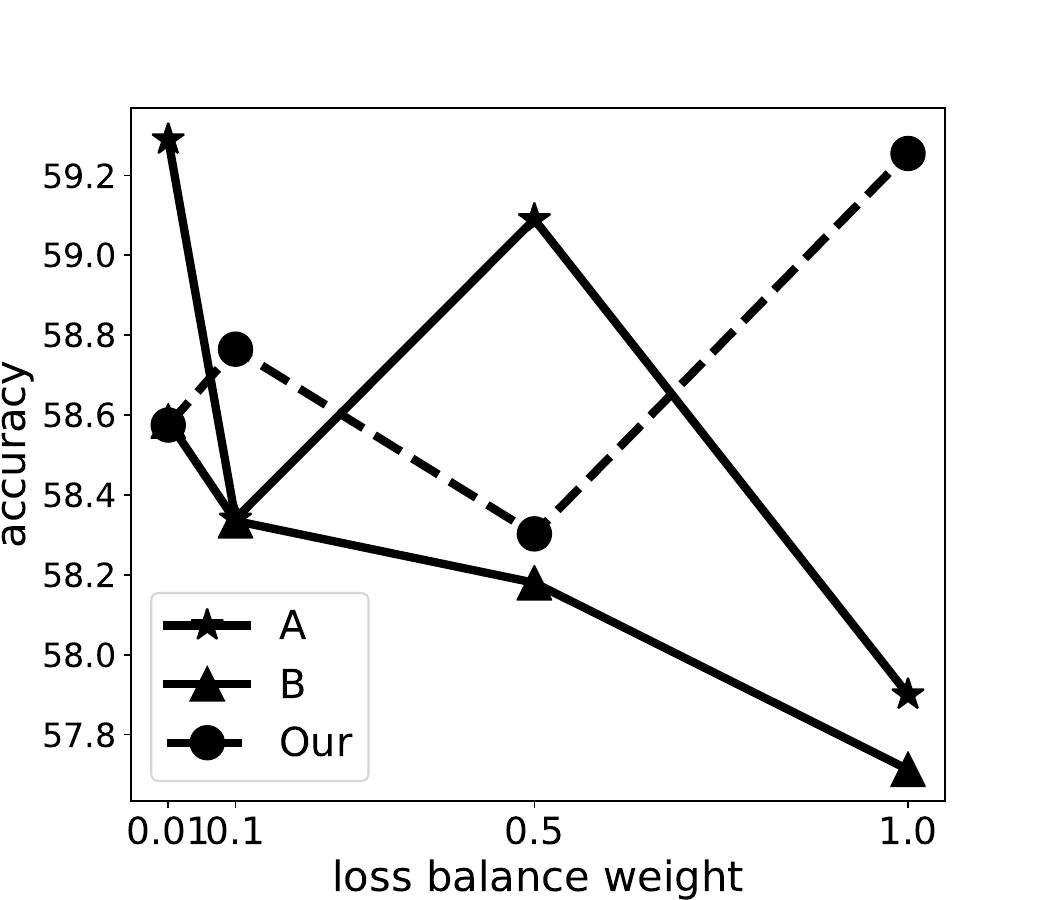}
\end{minipage}
\label{fig_c}
}
\subfigure[\imdbtwo]{
\begin{minipage}[t]{0.23\linewidth}
\centering
\includegraphics[width=1.6in]{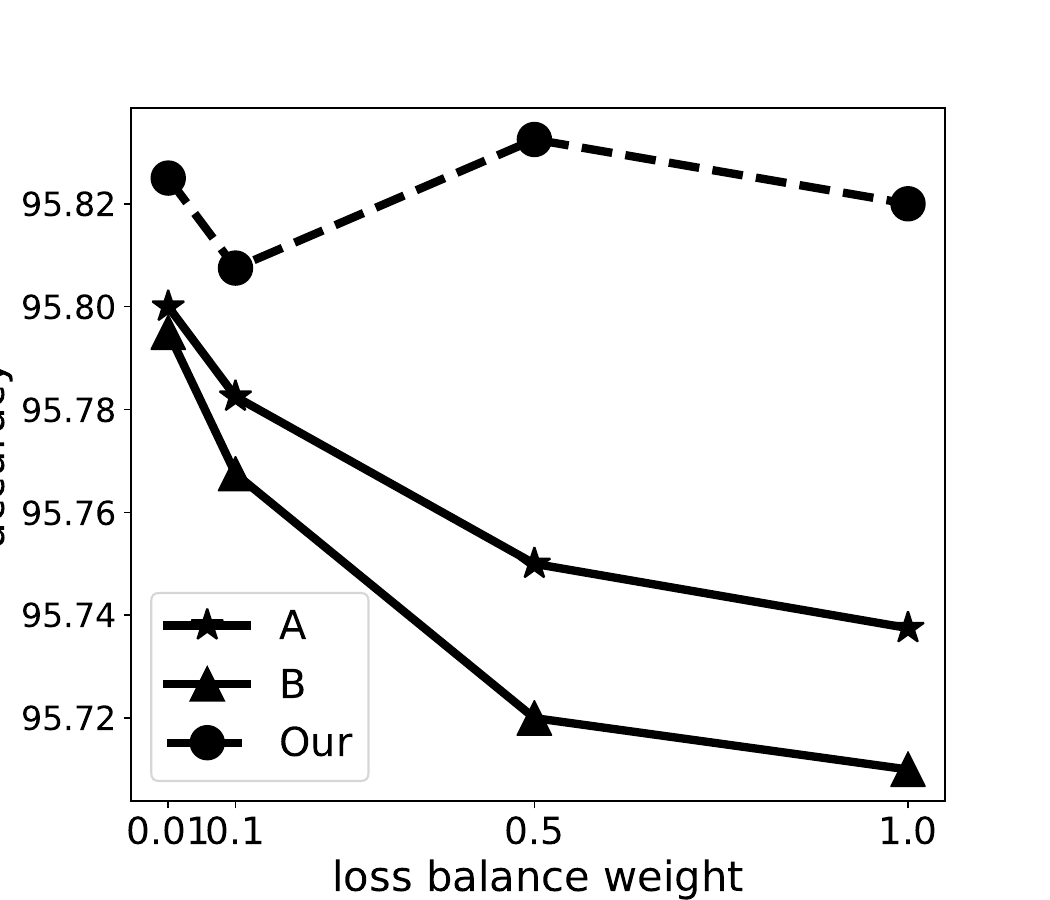}
\end{minipage}
\label{fig_d}
}
\centering
\caption{Impacts of loss balance weights, from left to right, are the results of \mr, \ssttwo, \sstfive~and \imdbtwo, respectively. A and B refer that auxiliary tasks A and B are tested solely. ``Our'' refers to \ourmodel.}
\label{figure_loss_balance_weight}
\end{figure*}

There are many alternatives to Equation~\ref{total_loss} for combining the losses. Previous work on multiple losses used only the sum~\cite{ke-etal-2020-sentilare}. The choice of the loss balance weight $\gamma$ is also important, as large values such as $\gamma = 1.0$ effectively reduce the weighting function to a simple sum over the losses, while smaller values (e.g., $\gamma$ = 0.01) allow the loss weights to vary.
Therefore, we search the loss balance weight $\gamma$ from $\{0.01, 0.1, 0.5, 1.0\}$ considering the following detailed considerations. First, we argue that higher auxiliary task weights may dominate the total loss, while smaller weights should be better, and 0.01 is selected. Second, the weights in $\left(0.0, 1.0\right]$ should be tested evenly. 
Figure~\ref{figure_loss_balance_weight} compares these alternatives, including auxiliary task SWS and CSP, and \ourmodel. It can be observed that, lower loss balance weight generally achieves better performance across most cases. Taking \imdbtwo~as an example, as there are more training samples and longer sequence length (512), making it less sensitive to seeds, with the decrease of loss balance weight, the advantages gradually increase, indicating that the weight of auxiliary tasks should be a small value to avoid undue impacts on the main task. 
Although for \mr, a dataset with a smaller training set, the results are sensitive to $\gamma$, a small $\gamma$ is also preferred in most cases.

\subsection{Analysis on Label Combination}

\begin{table}[t]
\centering
\resizebox{0.93\columnwidth}{!}{
\begin{tabular}{l|l|l|l|l}
\hline \textbf{Model}&   \textbf{\mr}&   \textbf{\ssttwo}&   \textbf{\imdbtwo}&  \textbf{\sstfive} \\ \hline
\sentix$_{\text{A+JC}}$&  87.31 & 92.20&    94.70&   55.74\\
\sentix$_{\text{A+CC}}$&  \textbf{87.35} &    \textbf{92.26}& \textbf{94.71}&   \textbf{55.81}  \\
\hline
\sentix$_{\text{B+JC}}$& 87.35 &  92.24&   94.59&  \textbf{55.83} \\
\sentix$_{\text{B+CC}}$& \textbf{87.38} & \textbf{92.59}& \textbf{94.61}&  55.74\\
\hline
\hline
\sentilare$_{\text{A+JC}}$& 90.69 & 94.72& 95.80& \textbf{59.29}  \\
\sentilare$_{\text{A+CC}}$& \textbf{90.74} & \textbf{94.91}&  \textbf{95.83}& 59.21 \\
\hline
\sentilare$_{\text{B+JC}}$& 90.88 & 94.91&  95.80& 58.59  \\
\sentilare$_{\text{B+CC}}$&  \textbf{91.10}& \textbf{94.99}& \textbf{95.84}& \textbf{58.97} \\
\hline
\end{tabular}
}
\caption{Comparison of joint combination (JC) and conditional combination (CC) in task A and B.}
\label{results_JLC} 
\end{table}

In addition to the auxiliary tasks, \ourmodel~also contains a label combination method unifying two different categories of labels (e.g., word/sentence sentiment label). To analyze the relative contribution of the conditional combination method compared to the joint combination method, we run additional comparison experiments that replace the joint combination with just the conditional combination method.
Table~\ref{results_JLC} summarizes the results for all evaluated datasets (\sentix~and \sentilare~are selected, as they perform better). Replacing the joint combination with the conditional combination gives a slight improvement for datasets \mr, \ssttwo~and \imdbtwo. For dataset \sstfive, the conditional combination is better than joint combination in some cases (e.g., from 58.59 accuracy to 58.97 for \sstfive~on the auxiliary task B). Overall the improvements are small compared to the full \ourmodel~model.
Joint combination is adopted by default in our experiments, as it is slightly easier to implement.

\subsection{Introduced Parameters}
For SWS, the number of increased parameters is $W_{\{i, j\}} \in \mathbb{R}^{|Y|d \times |C||Y|}, b_{\{i, j\}} \in \mathbb{R}^{|C||Y|}$ (Section~\ref{subtask_A}), sentiment word embedding table $E \in \mathbb{R}^{|V_1| \times d}$ and polarity embedding table $E_p \in \mathbb{R}^{|Z| \times d}$.
For CSP, the number of extra parameters is $W_3 \in \mathbb{R}^{d \times |Z||C|}, b_3 \in \mathbb{R}^{|Z||C|}$, sentiment word embedding table $E \in \mathbb{R}^{|V_1| \times d}$ and polarity embedding table $E_p \in \mathbb{R}^{|Z| \times d}$. 
The number of increased parameters induced by combining the two tasks is $W_4 \in \mathbb{R}^{2 \times 1}, b_4 \in \mathbb{R}$.
Therefore, the total number of parameters induced by \ourmodel~is $W_i, W_j, W_3, W_4, b_i, b_j, b_3, b_4$ and $E, E_p$, where $E, E_p$ is optional since it can be cached (just like GloVe~\citep{pennington2014glove}) and kept frozen to avoid introducing much parameters when the sentiment word vocabulary is large. In our experiments, $|C| \le 5$, $|Y|=|Z|=2$, $d=768$, $V_1=25,158$.

\section{Conclusion}


In this paper, we directly integrate sentiment knowledge into the fine-tuning phase. We design two sentiment-aware auxiliary tasks, SWS and CSP. 
SWS needs to select the correct sentiment words from the given options, while CSP predicts the overall sentiment with the word sentiment given as prior knowledge.
Further, we propose joint and conditional label combination methods to unify considered multiple kinds of labels into a single label. 
Though straightforward and conceptually simple, experiments demonstrate that \ourmodel~still further improves over solid baselines, verifying that \ourmodel~is complementary to existing sentiment-enhanced PLMs.

\section{Acknowledgement}
This work is supported in part by NSFC (61925203).

\bibliographystyle{acl_natbib}
\bibliography{anthology}

\end{document}